\documentclass[11pt,a4paper]{article}
\usepackage[hyperref]{acl2021}
\usepackage{times}
\usepackage{latexsym}

\usepackage{booktabs}
\usepackage{amssymb}
\usepackage{hyperref}
\usepackage{amsmath}

\usepackage{multirow}
\usepackage{latexsym}
\usepackage{booktabs}
\usepackage{courier}
\usepackage{amsmath}
\usepackage{subfig}
\usepackage{placeins}
\usepackage{tikz}
\usepackage{hyperref}

\usepackage{microtype}

\aclfinalcopy 

\title{Do Natural Language Explanations Represent Valid Logical Arguments? Verifying Entailment in Explainable NLI Gold Standards}

\author{Marco Valentino$^{\dagger}$, Ian Pratt-Hartmann$^{\dagger}$, Andr\'e Freitas$^{\dagger}$$^{\ddagger}$ \\  Department of Computer Science, University of Manchester, United Kingdom$^{\dagger}$ \\  Idiap Research Institute, Switzerland$^{\ddagger}$ \\ {\tt \{marco.valentino,ian.pratt,andre.freitas\}} \\ {\tt @manchester.ac.uk} \\}

\begin{document}
\maketitle
\begin{abstract}
An emerging line of research in Explainable NLP is the creation of datasets enriched with human-annotated explanations and rationales, used to build and evaluate models with step-wise inference and explanation generation capabilities. While human-annotated explanations are used as ground-truth for the inference, there is a lack of systematic assessment of their consistency and rigour.
In an attempt to provide a critical quality assessment of Explanation Gold Standards (XGSs) for NLI, we propose a systematic annotation methodology, named \emph{Explanation Entailment Verification} ($EEV$), to quantify the logical validity of human-annotated explanations.

The application of $EEV$ on three mainstream datasets reveals the surprising conclusion that a majority of the explanations, while appearing coherent on the surface, represent logically invalid arguments, ranging from being incomplete to containing clearly identifiable logical errors.
This conclusion confirms that the inferential properties of explanations are still poorly formalised and understood, and that additional work on this line of research is necessary to improve the way Explanation Gold Standards are constructed.
\end{abstract}

\section{Introduction}

Explanation Gold Standards (XGSs) are emerging as a fundamental enabling tool for step-wise and explainable Natural Language Inference (NLI). Resources such as WorldTree \cite{xie2020worldtree,jansen2018worldtree}, QASC \cite{khot2020qasc}, among others \cite{wiegreffe2021teach,thayaparan2020survey,bhagavatula2019abductive,camburu2018snli} provide a corpus of linguistic evidence on how humans construct explanations that are perceived as plausible, coherent and complete.

Designed for tasks such as Textual Entailment (TE) and Question Answering (QA), these reference datasets are used to build and evaluate models with step-wise inference and explanation generation capabilities \cite{valentino2021unification,cartuyvels2020autoregressive,kumar2020nile,rajani2019explain}. While these explanations are used as ground-truth for the inference, there is a lack of systematic assessment of their consistency and rigour, introducing inconsistency biases within the models.

\begin{figure}[t]
\centering
\includegraphics[width=0.9\columnwidth]{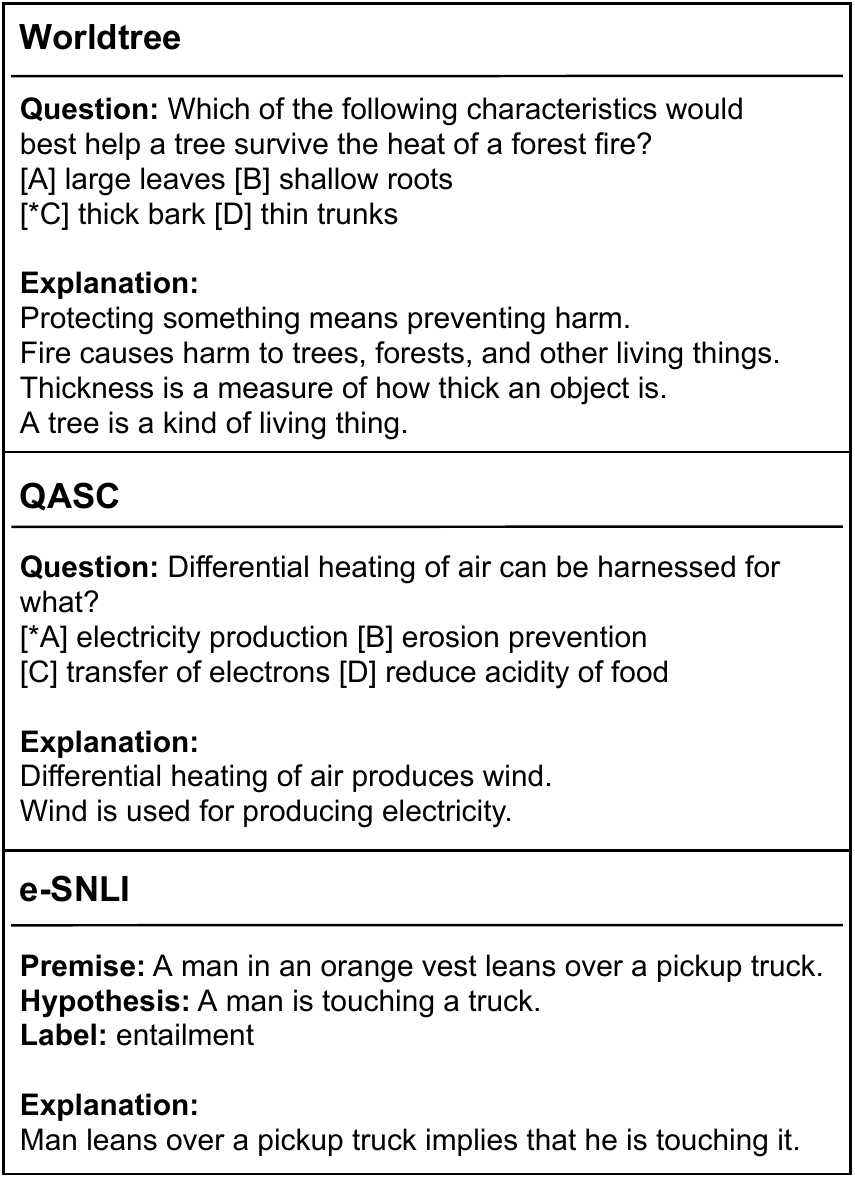}
\caption{Does the answer logically follow from the explanation? While step-wise explanations are used as ground-truth for the inference, there is a lack of assessment of their consistency and rigour. We propose $EEV$, a methodology to quantify the logical validity of human-annotated explanations.}
\label{fig:examples}
\end{figure}

This paper aims to provide a critical quality assessment of Eplanation Gold Standards for NLI in terms of their logical inference properties. By systematically translating natural language explanations into corresponding logical forms, we induce a set of recurring logical violations which can then be used as testing conditions for quantifying quality and logical consistency in the annotated explanations. 
More fundamentally, the paper reveals the surprising conclusion that a majority of the explanations present in explanation gold standards contain one or more major logical fallacies, while appearing to be coherent on the surface. This study reveals that the inferential properties of explanations are still poorly formalised and understood.

The main contributions of this paper can be summarised as:

\begin{enumerate}
    \item Proposal of a systematic methodology, named \emph{Explanation Entailment Verification} ($EEV$), for analysing the logical consistency of NLI explanation gold-standards.
    
    
    \item Validation of the quality assessment methodology for three contemporary and mainstream reference XGSs.
    
    \item The conclusion that most of the annotated human-explanations in the analysed samples represent logically invalid arguments, ranging from being incomplete to containing clearly identifiable logical errors.
    
\end{enumerate}

\section{Related Work}

An emerging line of research in Explainable NLP is focused on the creation of datasets enriched with human-annotated explanations and rationales \cite{wiegreffe2021teach}. These resources are often adopted as Explanation Gold Standards (XGSs), providing additional supervision for training and evaluating explainable models capable of generating natural language explanations in support of their predictions \cite{valentino2021unification,valentino2020explainable,kumar2020nile,cartuyvels2020autoregressive,thayaparan2020explanationlp,rajani2019explain}.

XGSs are designed to support Natural Language Inference, asking human-annotators to transcribe the reasoning required for deriving the correct prediction \cite{thayaparan2020survey}. Despite the popularity of these datasets, and their application for measuring explainability on tasks such as Textual Entailment \cite{camburu2018snli}, Multiple-choice Question Answering \cite{xie2020worldtree,jhamtani2020learning,khot2020qasc,jansen2018worldtree}, and other inference tasks \cite{wang2020semeval,ferreira2020premise,ferreira2020natural,bhagavatula2019abductive}, little has been done to provide a clear understanding on the nature and the quality of the reasoning encoded in the explanations.

Previous work on explainability evaluation has mainly focused on methods for assessing the quality and faithfulness of explanations generated by deep learning models \cite{camburu2020make,subramanian2020obtaining,kumar2020nile,jain2019attention,wiegreffe2019attention}. Our work is related to this research, but focuses instead on the resources on which explainable models are trained. In that sense, this paper is more aligned to gold standard evaluation methods, which aim to design systematic approaches to qualify the content and the inference capabilities involved in mainstream NLP benchmarks \cite{lewis-etal-2021-question,bowman2021will,schlegel2020framework,ribeiro2020beyond,pavlick2019inherent,min2019compositional}. However, to the best of our knowledge, none of these methods have been adopted to provide a critical assessment of human-annotated explanations present in XGSs.

\begin{figure*}[t]
\centering
\includegraphics[width=0.9\textwidth]{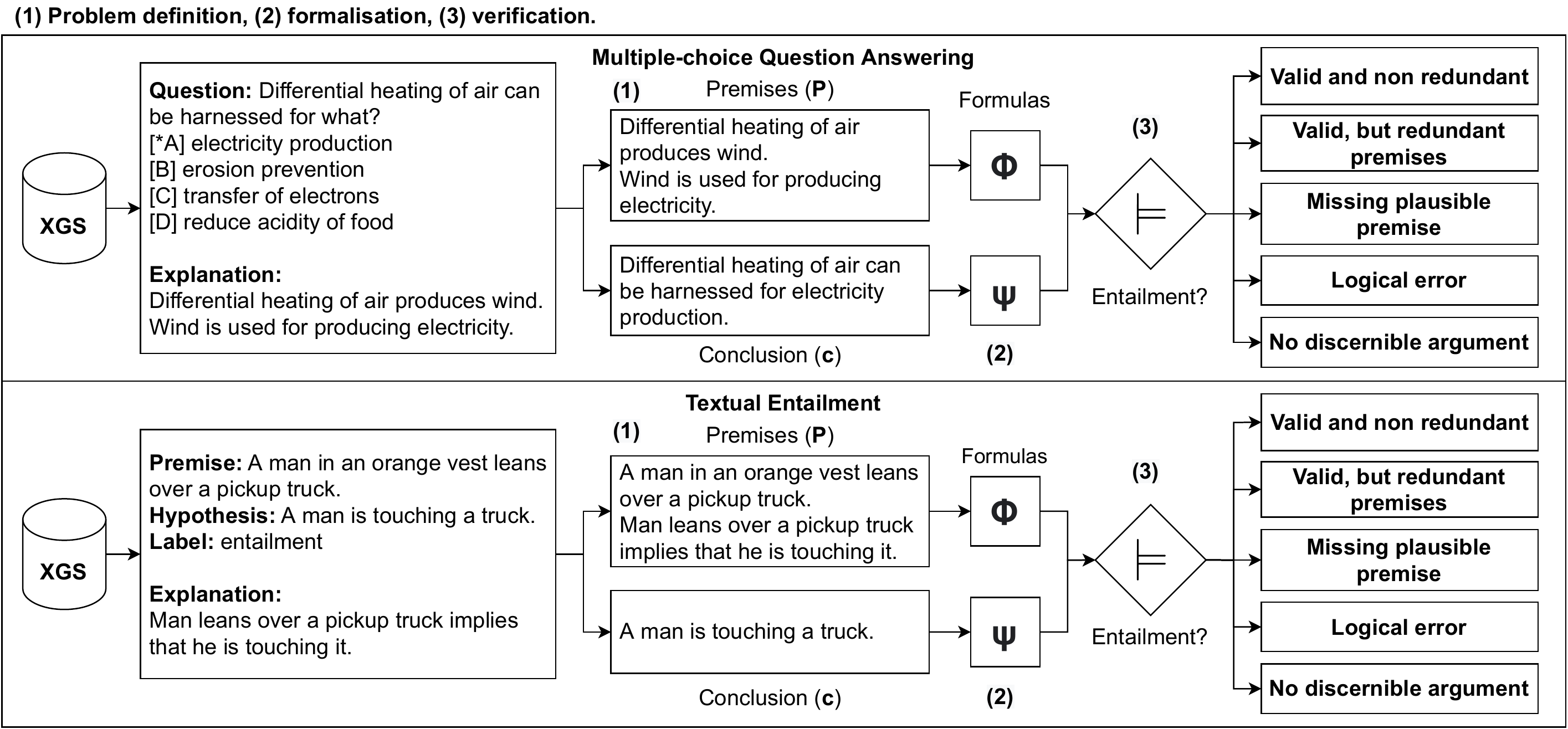}
\caption{Overview of the Explanation Entailment Verification ($EEV$) applied to different NLI problems. $EEV$ takes the form of a multi-label classification problem where, for a given NLI problem, a human annotator has to qualify the validity of the inference process described in the explanation through a pre-defined set of classes.}
\label{fig:methodology}
\end{figure*}

\section{Explanation Gold Standards}

Given a generic classification task $T$, an Explanation Gold Standard (XGS) is a collection of distinct instances of $T$, $XGS(T) = \{ I_1, I_2, \dots, I_n\}$, where each element of the set, $I_i = \{X_i, s_i, E_i\}$, includes a problem formulation $X_i$, the expected solution $s_i$ for $X_i$, and a human-annotated explanation $E_i$. 

In general, the nature of the elements in a XGS can vary greatly according to the task $T$ under consideration. In this work, we restrict our investigation to Natural Language Inference (NLI) tasks, such as Textual Entailment and Question Answering, where problem formulation, expected solution, and explanations are entirely expressed in natural language. 

For this class of problems, the explanation is typically a composition of sentences, whose role is to describe the reasoning required to arrive at the final solution. As shown in the examples depicted in Figure \ref{fig:examples}, the explanations are constructed by human annotators transcribing the commonsense and world knowledge necessary for the correct answer to hold.
Given the nature of XGSs for NLI, we hypothesise that a human-annotated explanation represents a valid set of premises from which the expected solution logically follows. 

In order to validate or reject this hypothesis, we design a methodology aimed at evaluating XGSs in terms of logical entailment, quantifying the extent to which human-annotated explanations actually entail the final answer.

\section{Explanation Entailment Verification}

We present a methodology, named Explanation Entailment Verification ($EEV$), aimed at quantifying and assessing the quality of human-annotated explanations in XGS for NLI tasks,  in terms of their logical inference properties.  

To this end, we design an annotation framework that takes the form of a multi-label classification problem defined on a XGS. Specifically, the goal of $EEV$ is to label each element in a XGS, $I_i = \{X_i, s_i, E_i\}$, using one of a predefined set of classes qualifying the validity of the inference process described in the explanation $E_i$. 

Figure \ref{fig:methodology} shows a schematic representation of the annotation pipeline. One of the challenges involved in the design of a standardised methodology for $EEV$ is the formalisation of an annotation task that is applicable to NLI problems with different shapes, such as Textual Entailment (TE) and Multiple-choice Question Answering (MCQA).  
To minimise the ambiguity in the annotation and make it independent of the specific NLI task, we define a methodology composed of three major steps: (1) \emph{problem definition}; (2) \emph{formalisation}; and (3) \emph{verification}.

In the problem definition step, each example $I_i$ in the XGS is translated into an entailment form ($P \models c$), identifying a set of sentences $P$ representing the premises for the entailment, and a single sentence  $c$ representing its conclusion. As illustrated in Figure \ref{fig:methodology}, this step defines an entailment problem with a single surface form that allows abstracting from the NLI task under investigation. 

In the formalisation step, the sentences in $P$ and $c$ are translated into a logical form ($\Phi \models \psi$). Specifically, the formalisation is performed using event-based semantics, in which verbs correspond to event-types, and their objects to semantic roles (additional details on the formalism are provided in section \ref{sec:formalisation}). This step aims to minimise the ambiguity in the interpretation of the meaning of the sentences, supporting the annotators in the identification of logical errors and gaps in the explanations, and maximise the inter-annotator agreement in the downstream verification task. 

The final step corresponds to the actual multi-label classification problem. Specifically, the annotators are asked to verify whether the formalised set of premises $\Phi$ entails the conclusion $\psi$ ($\Phi \models \psi$) and to classify the explanation in the corresponding example $I_i = \{X_i, s_i, E_i\}$ selecting one of the following classes: (1) \emph{Valid and non redundant}; (2) \emph{Valid, but redundant premises}; (3) \emph{Missing plausible premise}; (4) \emph{Logical error}; (5) \emph{No discernible argument}. The classes are mutually exclusive: each example can be assigned to one and only one label.  

After $EEV$ is performed for each instance in the dataset, the frequencies of the classification labels can be adopted to estimate and evaluate the overall entailment properties of the explanations in the XGS under consideration.

\subsection{Problem definition}

The problem definition step consists in the identification of the sentences in $I_i = \{X_i, s_i, E_i\}$ that will compose the set of premises $P$ and the conclusion $c$ for the entailment problem $P \models c$. 

Here, we describe the procedure adopted for translating a specific NLI task into the entailment problem of interest given its original surface form. In particular, we employ two different translation procedures for Textual Entailment (TE) and Multiple-choice Question Answering (MCQA) problems.

\paragraph{Textual Entailment (TE). }
For a TE task, the problem formulation $X_i$ is generally composed of two sentences, $p$ and $h$, representing a premise and a hypothesis (see e-SNLI in figure \ref{fig:examples}). Each example in a TE task can be classified using one of the following labels: \emph{entailment}, \emph{neutral}, and \emph{contradiction} \cite{bowman2015large}. 
In this work, we focus on examples where the expected solution $s_i$ is \emph{entailment}, implying that the hypothesis $h$ is a consequence of the premise $p$. Therefore, to define the entailment verification problem, we simply include the premise $p$ in $P$ and consider the hypothesis $h$ as a the conclusion $c$. 
For this class of problems, the explanation $E_i$ describes additional factual knowledge necessary for the entailment $p \models h$ to hold \cite{camburu2018snli}. Specifically, the sentences in $E_i$ can be interpreted as a further set of premises for the entailment verification problem and are included in $P$.

\paragraph{Multiple-choice Question Answering (MCQA). } In the case of MCQA, $X_i$ is typically composed of a question $Q_i = \{c_1, \ldots, c_n, q\}$, and a set of mutually exclusive candidate answers $A_i = \{a_1, \ldots, a_m\}$  (see QASC and Worldtree in figure \ref{fig:examples}). In this case, the expected label $s_i$ corresponds to one of the candidate answers in $A_i$ \cite{jansen2018worldtree,khot2020qasc}. 
$Q_i$ can include a set of introductory sentences $c_1, \ldots, c_n$ acting as a context for the question $q$.
We consider each sentence $c_i$ in the context as a premise for $q$ and include it in $P$.
Similarly to TE, we interpret the explanation $E_i$ for a MCQA example as a set of premises that entails the correct answer $s_i$. Therefore, the sentences in $E_i$ are included in $P$. The question $q$ takes the form of an elliptical assertion, and the candidate answers are possible substitutions for the ellipsis. Therefore, to derive the conclusion $c$, we adopt the correct answer $s_i$ as a substitution for the ellipsis in $q$. Details on the formalisation adopted for MCQA problems are described in section \ref{sec:formalisation}.

\subsection{Verification} 

In the verification step, the annotators adopt the formalised set of premises $\Phi$ and conclusion $\psi$ to classify the entailment problem in one of the following categories:
\begin{enumerate}
\item \textbf{Valid and non-redundant: } The argument is formally valid, and all premises are required for the derivation.  
\item \textbf{Valid, but redundant premises: } The argument is formally valid, but some premises are not required for the derivation. This includes the cases where more than one premise is present, and the conclusion simply repeats one of the premises.  
\item  \textbf{Missing plausible premise: } The argument is formally invalid, but would become valid on addition of a reasonable premise, such as, for example, 
\emph{``If x affects y, then a change to x affects y''}, or \emph{``If x is the same height as y and y is not as tall as z then x is not as tall as z''}. 
\item \textbf{Logical error: } The argument is formally invalid, apparently as a result of confusing \emph{``and''} and \emph{``or''} or \emph{``some''} and \emph{``all''}, or of illicitly changing the direction of an implication.
\item \textbf{No discernible argument: } The argument is invalid, no obvious rescue exists in the form of a missing premise, and no simple logical error
can be identified.
\end{enumerate}

\subsection{Formalisation} 
\label{sec:formalisation}

In this section, we describe an example of formalisation for a MCQA problem. 
A typical multiple-choice problem is a triple consisting of a \textit{question} $Q$ together with a set of \textit{candidate answers} $A_1, \dots, A_m$. It is understood that $Q$ takes the form of a elliptical assertion, and the candidate answers are possible substitutions for the ellipsis. The task is to determine which of the candidate answers would result in an assertion entailed by some putative knowledge-base. The corpora investigated feature a list of multiple-choice textual entailment problems
together, in each case, with a specification of a correct answer and an {\em explanation} in the form of a set of assertions $\Phi$ from the knowledge base providing a justification for the answer. For example, the following problem together with its resolution is taken from the Worldtree corpus \cite{jansen2018worldtree}.
\paragraph{Question:}
A group of students are studying bean plants. All of the following traits are affected by changes in the environment except \dots
\paragraph{Candidate answers:} [A] leaf color. [B] seed type. [C] bean production. [D] plant height.
\paragraph{Correct answer:} B
\paragraph{Explanation:} (i) The type of seed of a plant is an inherited characteristic; (ii) Inherited characteristics are the opposite of learned characteristics; acquired characteristics; (iii) An organism's environment affects that organism's acquired characteristics; (iv) A plant is a kind of organism; (v) A bean plant is a kind of plant; (vi) Trait is synonymous with characteristic.

\begin{table*}[t]
\small
 \centering
 \begin{tabular}{p{4cm}|ccc}
      \toprule
        \textbf{Feature} & \textbf{Worldtree} & \textbf{QASC} & \textbf{e-SNLI}\\
         \midrule
            Task & MCQA & MCQA & TE\\
            Multi-hop & yes & yes & no\\
            Crowd-sourced & no & yes & yes\\
            Explanation type & generated + composed & composed & generated\\
            Avg. number of sentences & 6 & 2 & 1\\
    \bottomrule
 \end{tabular}
 \caption{Features of the datasets selected for the Explanation Entailment Verification ($EEV$). }
 \label{tab:datasets}
\end{table*}

In formalising such problems, we represent the question as a sentence of first-order logic featuring a schematic formula variable $P$ (corresponding to the ellipsis), and
the candidate answers as first-order formulas. In the above example, 
we assume that the essential force of the question to find a characteristic of plants {\em not} affected by those plants' environments. That is, we are asked for a $P$
making the schematic formula
\begin{multline}
\forall x y z  w e  (\mbox{bnPlnt}(x) \wedge \mbox{env}(y,x) \wedge  \hspace{-2mm} \\ 
\mbox{changeIn}(z,y) \wedge \mbox{trait}(w,x) \wedge \mbox{affct}(e) \wedge\\
\hspace{-4mm}  \mbox{agnt}(e,z) \wedge P \rightarrow \neg \mbox{ptnt}(e,w)).
\label{eq:qn}
\end{multline}
into a true statement. We formalise the correct answer (B) by the atomic formula $\mbox{sdTp}(w,x)$ ``$w$ is the seed type of $x$'', with the other candidate answers
formalised similarly. In choosing predicates for formalisation,
we typically render common noun-phrases using predicates, taking these to be relational if the context demands (e.g.~``environment/seed type {\em of} a plant $x$'').
In addition, we typically render verbs as predicates whose arguments range over eventualities (events, processes, etc.), related to their participants via a
standard list of binary ``semantic role'' predicates (agent, patient, theme) etc. Thus, to say that ``$x$ affects $y$'' is to report the existence of
an eventuality $e$ of type ``affecting'', such that $x$ is the agent of $e$ and $y$ its patient. This approach, although somewhat strained
in many general contexts, 
aids standardization and, more importantly, also makes it easier to deal with adverbial phrases. Of course, many choices in formalisation strategy inevitably remain.

The knowledge-base excerpt $\Phi$ is formalised straightforwardly as a finite set of first-order formulas, following the same general rendering policies. In the case of the 
above example, sentences (i), (ii) and (iv)--(vi) in $\Phi$ might be formalised as:
\begin{align*}
&
\forall x y (\mbox{plnt}(x) \hspace{-1mm} \wedge \hspace{-1mm} \mbox{sdTp}(y,x) \rightarrow \mbox{char}(y,x) \hspace{-1mm} \wedge \hspace{-1mm}  \mbox{inhtd}(y))\\
& \forall x y (\mbox{char}(x,y) \wedge \mbox{inhtd}(x) \rightarrow \neg \mbox{acqrd}(x))\\
&\forall x (\mbox{plnt}(x) \rightarrow \mbox{orgnsm}(x))\\
&\forall x (\mbox{bnPlnt}(x) \rightarrow \mbox{plnt}(x))\\
&\forall x y (\mbox{trait}(x,y) \leftrightarrow \mbox{char}(x,y)),
\end{align*}
with the more complicated sentence (iii) formalised as 
\begin{align}
\begin{split}
& \forall xyw(\mbox{orgnsm}(x) \wedge \mbox{env}(y,x) \wedge\\
& \quad \mbox{char}(w,x) \wedge \mbox{acqrd}(w) \rightarrow\\
& \qquad \exists e (\mbox{affct}(e) \wedge \mbox{agnt}(e,y) \wedge  \mbox{ptnt}(e,w)))
\end{split}
\label{eq:badpremise3}
\end{align}
Denoting by $\psi$ the result of substituting $\mbox{sdTp}(w,x)$ for $P$ in~\eqref{eq:qn}, we ask ourselves: Does $\Phi$ entail $\psi$? A moment's
thought shows that it does not. At the very least, statement (iii) in the explanation,
whose {\em prima facie} formalisation is~\eqref{eq:badpremise3}, 
must instead be read as asserting that 
an organism's environment affects {\em only} that organism's acquired characteristics, 
that is to say:
\begin{align}
\begin{split}
& \forall x y w (\mbox{orgnsm}(x) \wedge \mbox{env}(y,x) \wedge \mbox{char}(w,x) \wedge \hspace{-2mm} \\ 
& \ \exists e (\mbox{affct}(e) \wedge \mbox{agnt}(e,y) \wedge \mbox{ptnt}(e,w)) \rightarrow \\
& \qquad  \qquad  \mbox{acqrd}(w)).
\end{split}
\label{eq:goodpremise3}
\end{align}
This is not unreasonable, of course. Generalizations in natural language are notoriously vague as to the direction of implication;
let $\Phi'$ be the result of substituting~\eqref{eq:goodpremise3}
for~\eqref{eq:badpremise3} in $\Phi$. Does $\Phi'$ entail $\psi$? Again, no. The problem this time is that, model-theoretically
speaking, just because something is affected by a {\em change in} its environment, that does not mean to say it is affected by its environment.
An assertion to the effect that it is would have to be postulated:
\begin{align*}
\begin{split}
& \forall x y z w (\mbox{env}(y,x) \wedge \mbox{changeIn}(z,y) \wedge\\
& \quad \exists e (\mbox{affct}(e) \wedge \mbox{agnt}(e,z) \wedge \mbox{ptnt}(e,w)) \rightarrow\\
& \quad \exists e (\mbox{affct}(e) \wedge \mbox{agnt}(e,y) \wedge \mbox{ptnt}(e,w))).
\end{split}\\
\end{align*}
Let $\Phi''$ be the result of augmenting $\Phi'$ in this way. Then $\Phi''$ does indeed entail $\psi$. 

Applying a general principle of charity, it is reasonable to take the interpretation of the explanation to be given by $\Phi'$. However, 
the additional premise required to obtain $\Phi''$ seems to have been forgotten.
Although not a logical truth, it has the status of a plausible general principle of the kind that is frequently explicitly articulated in the Worldtree database.  Therefore, we classify this example as a \textit{missing plausible premise}.
 
\section{Corpus Analysis}

We employ $EEV$ to analyse a set of contemporary XGSs designed for Textual Entailment and Multiple-choice Question Answering. 

In the following sections, we describe the methodology adopted for extracting a  representative sample from the selected XGSs, and for implementing the annotation pipeline efficiently. Finally, we present the results of the annotation, reporting the frequency of each entailment verification class and presenting a list of qualitative examples to provide additional insights on the logical properties of the analysed explanations.    

\subsection{Selected Datasets}

We select three contemporary XGSs with different and complementary characteristics. In particular, we apply our methodology to two MCQA datasets (Worldtree \cite{jansen2018worldtree}, QASC\cite{khot2020qasc}) and one TE benchmark (e-SNLI \cite{camburu2018snli}). 

The main features of the selected XGSs are reported in Table \ref{tab:datasets}. \emph{Multi-hop} indicates whether the problem requires step-wise reasoning, combining more than one sentence to compose the final explanation. \emph{Crowd-sourced} indicates whether the resource is curated using standard crowd-sourcing platforms. \emph{Explanation type} represents the methodology adopted to construct the explanations. \emph{Generated} means that the sentences in the explanations are entirely created by human annotators. On the other hand, \emph{composed} means that the sentences are retrieved from an external knowledge resource. Finally, the last row reports the \emph{average number of sentences} composing the explanations.

\subsection{Annotation Task}

The bottleneck of the annotation framework lies in the formalisation phase, which is generally time consuming and requires trained experts in the field. In order to make the application of $EEV$ efficient in practice, we extract a sub-set of $n = 100$ examples from each XGS (Worldtree, QASC, and e-SNLI).
To maximise the representativeness of the explanations in the subset, given a fixed size $n$, we combine a set of sampling methodologies with effect size analysis. The details of the sampling methodology are described in section \ref{sec:sampling} while the results are presented in section \ref{sec:results}. Code and data adopted for the experiments are available online \footnote{\url{https://github.com/ai-systems/explanation-entailment-verification/}}.

The extracted examples are randomly assigned to 2 annotators with an overlap of 20 instances to compute the inter-annotator agreement. All the annotators are active researchers in the field of Natural Language Processing and Computational Semantics. Table \ref{tab:annotator_agreement} reports the inter-annotator agreement achieved on each dataset separately. 
Overall, we observe an average of 72\% accuracy in the multi-label classification task, computed considering the percentage of overlaps between the final entailment verification classes chosen by the annotators.

\subsection{Sampling Methodology}
\label{sec:sampling}
 
To maximise the representativeness of the explanations for the subsequent annotation task, while analysing a fixed number $n$ of examples for each dataset, we combine a set of sampling methodologies with effect size analysis. In this section, we describe the sampling techniques adopted for each dataset.

A stratified sampling methodology has been adopted for the Worldtree corpus \cite{xie2020worldtree,jansen2018worldtree}. The stratified sampling consists in partitioning the dataset using a set of classes and performing random sampling from each class independently. This strategy guarantees that the same amount of examples is extracted from each class. The stratified technique requires the classes to be collectively exhaustive and mutually exclusive -- i.e, each example has to belong to one and only one class.
To apply stratified sampling on Worldtree, we consider the high-level topics introduced in \cite{xu2020multi}, which are used to classify each question in the dataset according to one of the following categories: Life, Earth, Forces, Materials, Energy, Scientific Inference, Celestial Objects, Safety, Other.
The same technique cannot be applied to e-SNLI \cite{camburu2018snli} and QASC \cite{khot2020qasc} since the examples in these datasets are not partitioned using any abstract set of classes. In this case, therefore, we use random sampling on the whole dataset to extract a fixed number $n$ of examples. 

Once a fixed number of examples $n$ is extracted from each dataset, we consider the annotated explanation sentences of each example to verify whether the extracted set of explanations is representative of the whole dataset. To perform this analysis, we assume the predicates in the explanation sentences to be the expression of the type of knowledge of the whole explanation. Therefore, we consider the extracted sample of explanations representative if the distribution of predicates in the sample is correlated with the same distribution in the whole dataset.
To this end, we compute the frequencies of the verbs appearing in the explanation sentences from the extracted sub-set and original dataset separately. Subsequently, we compare the frequencies in the sub-sample with the frequencies in the whole dataset computing a Pearson correlation coefficient. In this case, a coefficient greater than $.7$ indicates a strong correlation between the types of explanations in the sample and the types of explanations in the original dataset. After running the sampling for $t$ times independently, we select the subset of explanations for each dataset with the highest Pearson correlation coefficient. 
Table \ref{tab:categories_analysis__results} reports the Pearson correlation for the subsets adopted in our analysis with fixed sample size $n = 100$. 

\begin{table}[t]
     \centering
     \small
     \begin{tabular}{p{2cm}|c}
          \toprule
            \textbf{Dataset} & \textbf{Agreement Accuracy}\\
            \midrule
            \textbf{Worldtree} & .70\\
            \textbf{QASC} &  .70\\
            \textbf{e-SNLI} & .75\\
          \bottomrule
     \end{tabular}
     \caption{Inter-annotator
     agreement computed in terms of accuracy in the multi-label classification task considering the first annotator as a gold standard.}
     \label{tab:annotator_agreement}
 \end{table}
 
  \begin{table}[t]
     \centering
     \small
     \begin{tabular}{p{2cm}|cc}
          \toprule
            \textbf{Dataset} & \textbf{Correlation Coefficient}\\
            \midrule
            \textbf{Worldtree} & .964\\
            \textbf{QASC} & .958 \\
            \textbf{e-SNLI} & .987 \\
          \bottomrule
     \end{tabular}
     \caption{Effect size analysis of the samples extracted from each XGS for the downstream $EEV$ annotation.}
     \label{tab:categories_analysis__results}
 \end{table}

\begin{table*}[t]
     \centering
     \small
     \begin{tabular}{p{6cm}|ccc|c}
          \toprule
            \textbf{Entailment Verification Class} & \textbf{Worldtree} & \textbf{QASC} & \textbf{e-SNLI} & \textbf{AVG}\\
             \midrule
                Valid and non-redundant & 12.24 & 17.65 & \underline{\textbf{31.37}} & 20.42 \\
                Valid, but redundant premises & 26.53 & 7.84 & \underline{\textbf{31.37}} & 21.91\\
                \midrule
                Missing plausible premise & \underline{\textbf{38.78}} & 21.57 & 17.65 & \textbf{26.00}\\
                Logical error & 6.12 & \underline{17.65} & 9.80 & 11.19\\
                No discernible argument & 16.33 & \underline{\textbf{35.29}} & 9.80 & 20.47\\
                \midrule
                \midrule
                Valid argument & 38.77 & 25.49 & \underline{\textbf{62.74}} & 42.33\\
                Invalid argument & 61.23 & \textbf{\underline{74.51}} & 37.25 & \textbf{57.66}\\
        \bottomrule
     \end{tabular}
     \caption{Results of the application of $EEV$ for each entailment verification category.}
     \label{tab:explanations_results}
 \end{table*}

\subsection{Results}
\label{sec:results}

The quantitative analysis presented in this section aims to empirically assess the hypothesis that human-annotated explanations in XGSs constitute \textit{valid and non-redundant} logical arguments for the expected answers. 
We report the quantitative results of the explanation entailment verification in Table \ref{tab:explanations_results}. Specifically, the table reports the percentage of the frequency of each verification class in the analysed samples. The column \emph{AVG} reports the average for each class. 

Overall, we observe that the results of the annotation task tend to reject our research hypothesis, with an average of only 20.42\% of analysed explanations being classified as \emph{valid and non redundant} arguments. When considering also \emph{valid, but redundant} explanations (21.91\%), the average percentage of valid arguments reaches a total of 42.33\%. Therefore, we can conclude that the majority of the explanations represent invalid arguments (57.66\%). 

We observed that the majority of invalid arguments are classified as \emph{missing plausible premise}. This finding implies that a significant percentage of annotated explanations are incomplete arguments (26.00\%), that can be made valid on addition of a reasonable premise. We attribute this result to the tendency of human explainers to take for granted part of the world knowledge required in the explanation \cite{walton2004new}.

A lower but significant percentage of explanations contain identifiable logical errors (11.19\%), which result from confusing the set of quantifiers and logical operators, or from illicitly changing the direction of an implication. Similarly, 20.47\% of the explanations where labeled as \emph{no discernible arguments}, where no obvious premise can be added to make the argument valid and no simple logical error can be detected. This result can be attributed partly to natural errors occurring in a gold standard creation process, partly to the effort required for human-annotators to identify logical fallacies in their explanations. 
In the remaining of this section, we analyse the results obtained on each XGS.

\paragraph{Worldtree.} The analysed sample contains the highest percentage of incomplete arguments, with a total of 38.78\% explanations classified as \emph{missing plausible premise}. This result can be explained by the fact that the questions in Worldtree require complex forms of reasoning, facilitating the construction of arguments containing implicit world knowledge and missing premises. At the same time, the dataset contains the smallest percentage of logical errors (6.12\%). We attribute this outcome to the fact that Worldtree is not crowd-sourced, implying that the quality of the annotated explanations is more easily controllable using internal verification methods.

\paragraph{QASC.} This XGS contains the highest rate of invalid  arguments (62.74\%), with 35.29\% of the explanations classified as \emph{no discernible argument}. One of the factors contributing to these results might be related to the length of the constructed explanations, which is limited to 2 facts extracted from a predefined corpus of sentences. The high rate of no discernible arguments and missing premises (35.29\% and 21.57\% respectively) suggests that the majority of the questions require additional world knowledge and more detailed explanations. This conclusion is also supported by the percentage of \emph{valid, but redundant} arguments, which is the lowest among the analysed samples (7.84\%). Finally, the highest rate of logical errors (17.65\%) might be due to a combination of factors, including the complexity of the question answering task and the adopted crowd-sourcing mechanism, which prevent a thorough quality assessment.  

\paragraph{e-SNLI.} The sample includes the highest percentage of valid arguments with a total of 31.37\%. However, we noticed that the complexity of the reasoning involved in e-SNLI is generally lower than Worldtree and QASC, with most of the textual entailment problems being an example of \textit{monotonicity reasoning}. This observation is supported by the highest percentage of \emph{valid, but redundant} cases (31.37\%), where the explanation simply repeats the content of the conclusion. This occurrs quite often for examples of lexical entailment, where the words in the conclusion are a subset of the words in the premise. The lexical entailment instances, in fact, do not require any additional world knowledge, making any attempt of constructing an explanation redundant. Despite these characteristics, our evaluation suggests that a significant percentage of arguments are invalid (37.25\%). Again, this percentage might be the results of different factors, including the errors produced by the crowd-sourcing process.

  \begin{table*}[t]
    \centering
    \small
    \begin{tabular}{@{}p{7cm}p{7cm}c@{}}
    \toprule
         \textbf{Problem Formulation} &
         \textbf{Explanation} &
         \textbf{XGS} \\
         \midrule
         \textbf{Valid and non-redundant (20.42\%)}\\
         \midrule
         \textbf{Premise:} A smiling woman is playing the violin in front of a turquoise background. \textbf{Hypothesis:} A woman is playing an instrument.
         & A violin is an instrument. & e-SNLI\\
         \midrule
         \textbf{Valid, but redundant premises (21.91\%)}\\
         \midrule
        \textbf{Premise:} Four people are bandaging a head wound. \textbf{Hypothesis:} People are bandaging an injured head.
        & People are bandaging an injured head wound. & e-SNLI\\
         \midrule
         \textbf{Missing plausible premise (26.00\%)} \\
         \midrule
        \textbf{Question:} A group of students are studying bean plants.  All of the following traits are affected by changes in the environment except [A] Leaf color [*B] Seed type [C] Bean production [D] Plant height & The type of seed of a plant is an inherited characteristic. Inherited characteristics are the opposite of learned characteristics; acquired characteristics. An organism’s environment affects that organism’s acquired characteristics. A plant is a kind of organism. Trait is synonymous with characteristic. & Worldtree\\
         \midrule
         \textbf{Logical error (11.19\%)}\\
         \midrule
         \textbf{Question:} What group of animals do chordates belong to? [A] graptolites [B] more abundant [C] warm-blooded [D] four limbs [E] epidermal [*F] Vertebrates [G] animals [H] insects & Chordates have a complete digestive system and a closed circulatory system. Vertebrates have a closed circulatory system. & QASC\\
         \midrule
         \textbf{No discernible argument (20.47\%)}\\
         \midrule
          \textbf{Question:} What do plants require for reproduction? [A] energy [B] nutrients [C] bloom time [*D] animals [E] sunlight [F] Energy. [G] food [H] hormones & Plants require seed dispersal for reproduction.
          Seeds are probably dispersed by animals. & QASC \\
         \bottomrule
    \end{tabular}
    \caption{Examples of explanations classified with different entailment verification categories.}
    \label{tab:qualitative_examples}
\end{table*}

Table \ref{tab:qualitative_examples} reports a set of representative cases extracted from the evaluated samples. For each entailment verification class, we report an example extracted from the XGS with the highest percentage of instances in that class.  

 \begin{table}[t]
     \small
     \centering
     \begin{tabular}{cc}
          \toprule 
          \textbf{Dataset} & \textbf{Non contrastive explanations} \\
           \midrule
            Worldtree & 26.53\\ 
            QASC & \underline{\textbf{49.02}}\\
        \bottomrule
     \end{tabular}
     \caption{ Percentage of explanations in the MCQA sample labeled as non contrastive.}
     \label{tab:answers_result}
 \end{table}

\subsection{Contrastive Explanations}

Previous studies highlight the fact that explanations are \emph{contrastive} in nature, that is, they describe why an event \textit{P} happened instead of some counterfactual event \textit{Q} \cite{miller2019explanation,lipton1990contrastive}. Following this definition, we perform an additional analysis to verify whether the explanations contained in MCQA datasets are \emph{contrastive} with respect to the wrong candidate answers -- i.e., the explanation supports the validity of the correct answer while excluding the set of alternative choices. 
In order to quantify this aspect, we asked the annotators to label the questions with more than one plausible answer, whose explanations do not mention any discriminative commonsense or world knowledge that explains why the gold answer is correct instead of the alternative choices.

The results of this experiment are reported in Table \ref{tab:answers_result}. Overall, we found that a significant percentage of explanations are labeled as non contrastive. This outcome is particularly evident for QASC. We attribute these results to the presence of multi-adversary answer choices in QASC, which are generated automatically to make the dataset more challenging for language models. However, we found that this mechanism can produce questions with more than one plausible correct answer, which can cause the explanation to loose its contrastive function (see QASC examples in Table \ref{tab:qualitative_examples}). 

\section{Conclusion and Future Work}

This paper proposed a systematic annotation methodology to quantify the logical validity of human-annotated explanations in Explanation Gold Standards (XGSs).
The application of the framework on three mainstream datasets led us to the conclusion that a majority of the explanations represent logically invalid arguments, ranging from being incomplete to containing clearly identifiable logical errors. 

The main limitation of the framework lies in the scalability of its current implementation, which is generally time consuming and requires trained semanticists. One way to improve its efficiency is to explore the adoption of supporting tools for the formalisation, such as semantic parsers and/or automatic theorem provers.

Despite the current limitations, this study offers some important pointers for future work.
On the one hand, the results suggest that logical errors can be reduced by a careful design of the gold standard, such as authoring explanations with internal verification strategies or reducing the complexity of the reasoning task.
On the other hand, the finding that a large percentage of curated explanations still represent incomplete arguments has a deeper implication on the nature of explanations and on what annotators perceive as a valid and complete logical argument. Therefore, we argue that future progress on the design of XGSs will depend, among other things, on a better formalisation and understanding of the inferential properties of explanations.

\bibliographystyle{acl_natbib}
\bibliography{acl2021}


\end{document}